\documentclass[fleqn,10pt]{wlscirep}
\usepackage[utf8]{inputenc}
\usepackage[T1]{fontenc}
\usepackage{booktabs}
\usepackage{multirow}

\title{QuATON: \underline{Qu}antization \underline{A}ware \underline{T}raining of \underline{O}ptical \underline{N}eurons}

\author[1]{Hasindu Kariyawasam}
\author[1]{Ramith Hettiarachchi}
\author[2,3]{Quansan Yang}
\author[3]{Alex Matlock}
\author[2,4]{Takahiro Nambara}
\author[2,4]{Hiroyuki Kusaka}
\author[2,4]{Yuichiro Kunai}
\author[3,5]{Peter T C So}
\author[2,5,6,7,8,9,10]{Edward S Boyden}
\author[1,*]{Dushan Wadduwage}
\affil[1 ]{Center for Advanced Imaging, Faculty of Arts and Sciences, Harvard University, Cambridge, MA, USA}

\affil[2 ]{McGovern Institute for Brain Research, Massachusetts Institute of Technology (MIT), Cambridge, MA 02139, USA.}
\affil[3 ]{Department of Mechanical Engineering, MIT, Cambridge, MA 02139, USA.}
\affil[4 ]{Advanced Research Core, Fujikura Ltd., Kiba, Tokyo, Japan}
\affil[5 ]{Department of Biological Engineering, MIT, Cambridge, MA 02139, USA.}
\affil[6 ]{Department of Brain and Cognitive Sciences, MIT, Cambridge, MA 02139, USA.}
\affil[7 ]{Howard Hughes Medical Institute, MIT, Cambridge, MA 02139, USA.}
\affil[8 ]{K. Lisa Yang Center for Bionics, MIT, Cambridge, MA 02139, USA.}
\affil[9 ]{Center for Neurobiological Engineering, MIT, Cambridge, MA 02139, USA.}
\affil[10]{Koch Institute for Integrative Cancer Research, MIT, Cambridge, MA 02139, USA. \newline}

\affil[* ]{wadduwage@fas.harvard.edu}

\begin{abstract}

Optical processors, built with "optical neurons", can efficiently perform high-dimensional linear operations at the speed of light. Thus they are a promising avenue to accelerate large-scale linear computations. With the current advances in micro-fabrication, such optical processors can now be 3D fabricated, but with a limited precision. This limitation translates to quantization of learnable parameters in optical neurons, and should be handled during the design of the optical processor in order to avoid a model mismatch. Specifically, optical neurons should be trained or designed within the physical-constraints at a predefined quantized precision level. To address this critical issues we propose a physics-informed quantization-aware training framework. Our approach accounts for physical constraints during the training process, leading to robust designs. We demonstrate that our approach can design state of the art optical processors using diffractive networks for multiple physics based tasks despite quantized learnable parameters. We thus lay the foundation upon which improved optical processors may be 3D fabricated in the future. 

\end{abstract}
\begin{document}

\flushbottom
\maketitle
%
%
\thispagestyle{empty}


\section*{Introduction}
\label{sec:introduction}

Current advancements in deep learning have renewed the interest in optical processors as a promising avenue to accelerate large-scale linear computations. While such optical accelerators have not yet been fully realized for arbitrary linear operations, optical neural processors -i.e. optical processors equivalent to neural networks- have already shown great promise in a variety of applications. For instance, Diffractive networks ~\cite{Lin2018All-opticalNetworks}, made of a cascade of passive diffractive layers, could perform all-optical classification~\cite{Lin2018All-opticalNetworks}, all-optical quantitative phase imaging (QPI)~\cite{mengu2022all, herath2022differentiable}, optical logic operations~\cite{qian2020performing}, spatiotemporal signal processing~\cite{zhou2024spatiotemporal}, saliency segmentation~\cite{yan2019fourier}, and 3D object detection~\cite{shi2021multiple}. Similarly, learnable optical Fourier processors have shown to be capable of all-optical quantitative phase imaging~\cite{herath2022differentiable}, medical image processing~\cite{yelleswarapu2008optical, panchangam2001processing}, optical image encryption~\cite{liu2001optical}, and image classification~\cite{miscuglio2020massively}. Such optical processors have also been used as coding elements to design end-to-end optimized computational imaging systems~\cite{Haputhanthri2022FromMicroscopy,Kellman2019Physics-BasedImaging}. All these optical processors linearly map an input light field to an output field. They thus perform specific linear operations “learned” from training data, using computational elements we term “optical neurons”. 
\newline 

Optical neurons are traditionally implemented using reconfigurable optical elements like digital micromirror devices (DMDs), and spatial light modulators (SLMs). The state of each DMD micromirror or each SLM pixel, is treated as the learnable parameter of the optical neuron. However, recent advancements in microfabrication have now enabled the fabrication of 3D optics by voxel-wise modulating the refractive index of an optical substrate. Therefore we now possess the unique capability to 3D print optical processors upon learnable design. Here, the transmission coefficient at each 3D location is treated as the learnable parameter of the optical neuron. In either case, however, —unlike the parameters of artificial neurons that can represent any real value— the parameters of optical neurons can only represent a limited set of complex values, constrained due to their physical characteristics and fabrication limitations. More precisely, parameters in optical neurons are quantized, complex-valued, and bounded. For example, typical SLM pixels enable 8-bit-quantized phase-only parameters bounded in $[0, 2\pi)$;  DMD micromirrors enable 1-bit-quantized amplitude-only parameters bounded in $\{0, 1\}$. Similarly in 3D optics, fabrication limitations essentially constrain the parameters to a set of quantized complex values that are bounded in phase ($[0, 2\pi)$). These constrains may be ignored while designing (or training) the optical processor. But the change in the parameter distribution from design to fabrication, i.e., the model mismatch, frequently leads to a performance decline in the realized system~\cite{Metzler2020DeepImaging}. For example, recent work on all-optical QPI using diffractive networks (D2NNs) shows that reducing the precision of physical parameters exponentially decreases the performance of the D2NN~\cite{mengu2022all}. Nevertheless, beyond such isolated examples, there has been little exploration~\cite{Li2022Physics-AwareNetworks} on how both \textit{boundedness} and \textit{quantization} of parameters affect the performance of optical neurons.

\begin{figure*}
    \centering
    \includegraphics[width= 0.5 \linewidth ]{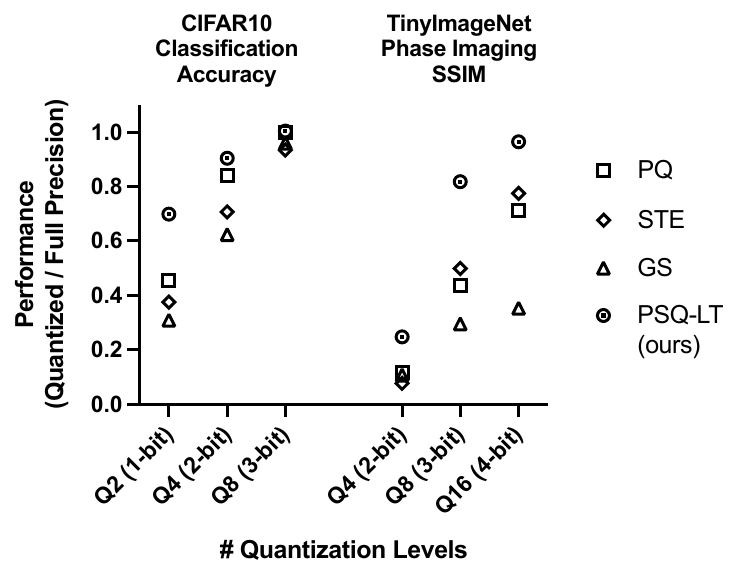}
    \caption{\textbf{Representative results for performance of QuATON (PSQ-LT) compared to other QAT methods (PQ, STE, GS) used to train optical neurons}. Here PSQ-LT, PQ, STE, and GS respectively stands for progressive sigmoid quantization with learnable temperature, post quantization, straight-through estimator, and gumbel softmax.}
    \label{fig:summary}
\end{figure*}

While quantization of optical neurons remain unexplored, quantization of parameters in artificial neural networks has been extensively studied in the machine learning literature to deploy models in resource constraint devices~\cite{zafrir2019q8bert, wu2016quantized,esser2019learned}. These works address, quantization post hoc~\cite{liu2021post, nagel2020up, nahshan2021loss} or during the training --using quantization aware training (QAT) methods~\cite{Yang2019QuantizationNetworks,esser2019learned, Gong2019DifferentiableNetworks}. Post-hoc-quantized models are easier to train but perform poorly during inference due to model mismatch; QAT models, on the other hand, are harder to train but perform faithful to the trained model. The training difficulty lies in the quantization operation, which is not differentiable. Non-differentiability impedes gradient-based optimizers making quantized models harder to train. QAT methods address this issue using few ways. Most straight forward is to inject quantization noise during forward propagation such that the model learns to be robust; this technique has been used in optical neural network to train neuromorphic models~\cite{kirtas2022quantization}. Another is to use the straight-through estimator (STE)~\cite{bengio2013estimating, gu2020roq} that passes the output gradient directly to the input during backpropagation; STE has been used to train optical neurons with limited-precision controls and device variations~\cite{gu2020roq}. A third approach is to use differentiable soft quantization functions~\cite{Gong2019DifferentiableNetworks, Yang2019QuantizationNetworks} that approximate the quantized model. Soft quantization functions have not been efficiently utilized in optical neurons, and is the focus of this work. Of note however, Gumbel-Softmax (GS) --that enjoys wide-spread use in neural network as a differentiable approximation to the categorical distribution-- has been used as QAT technique in optical neural networks~\cite{Li2022Physics-AwareNetworks}. While conceptually similar, GS is not a direct approximation of the quantization operation and hence is not as effective as soft quantization functions (see Fig.~\ref{fig:summary}). 
 


To this end, inspired by recent work in computer vision~\cite{Gong2019DifferentiableNetworks, Yang2019QuantizationNetworks}, we introduce "Quantization Aware Training of Optical Neurons", or "QuATON" a QAT framework specifically targeting optical neurons. The key elements of this framework consists of: (1) a soft quantization function constructed using shifted sigmoid functions that gradually evolve to the desired hard quantization levels; (2) an auto-tuning temperature parameter to control the quantization function during training. In comprehensive performance comparisons, we show the superiority of QuATON over competing methods used to train optical neurons (see Fig.~\ref{fig:summary}). Our work sets the foundation upon which improved optical processors may be designed and built in the future.

\section*{Results and Discussion}
\label{sec:results}

\subsection*{QuATON: Quantization-aware Training of Optical Neurons}

The quantization operation maps a given continuous variable, $x \in \mathbb{R}$, to a set of discrete values. These discrete values are known as quantization levels; In uniform quantization, they are evenly spaced and the uniform quantization operation is defined as,
 \begin{equation}
    \label{eq:uniform_quant}
    Q_h(x) = 
    \begin{cases}
        l & \text{if } x < l\\
        round\left(\frac{x-l}{\Delta}\right) \Delta + l & \text{if } l \leq x < u \\
        u & \text{if } x \geq u
    \end{cases}.
\end{equation}   
Here, $round(.)$ is the rounding operation to the nearest integer, and $\Delta = \frac{u-l}{N-1}$ is the step size. $l, u \in \mathbb{R}$ are the lower and upper bounds of the quantized range, where $N \in \mathbb{Z}$ is the number of quantization levels. We hereon refer to the operation described in Eq.~\ref{eq:uniform_quant} as the \emph{hard-quantization function}.
    
Due to its step-like nature, the hard-quantization function (Eq.~\ref{eq:uniform_quant}) has a zero derivative almost everywhere except for the sharp transitions between two quantization levels; at the transitions its derivative is undefined. Gradient-based optimizers --that depend on the derivative-- thus fail in the presence of the hard-quantization function. In artificial- or deep- neural networks, the issues of hard-quantization have been addressed using soft-quantization functions. For instance, Differentiable soft quantization (DSQ)~\cite{Gong2019DifferentiableNetworks} is a soft-quantization function based on the hyperbolic tangent (tanh). It has a learnable temperature parameter controlling the steepness of the transitions between two quantization levels. The temperature is updated during the training process which transforms the DSQ function closer to hard-quantization. Inspired by this, we developed a separate soft-quantization function named \emph{progressive sigmoid quantization (PSQ)} to train optical neurons aware of quantization. PSQ is different from DSQ in two ways. First, PSQ is based on the sigmoid function. Second, in DSQ, the weights outside the quantization range are clamped to the lower and upper bounds of the range. This results in a zero gradient for the weights outside the quantization range, causing them to freeze during training. In PSQ, we removed this clamping which gives a non-zero gradient to weights outside the quantization range. 

\begin{figure*}
    \centering
    \includegraphics[width=\linewidth]{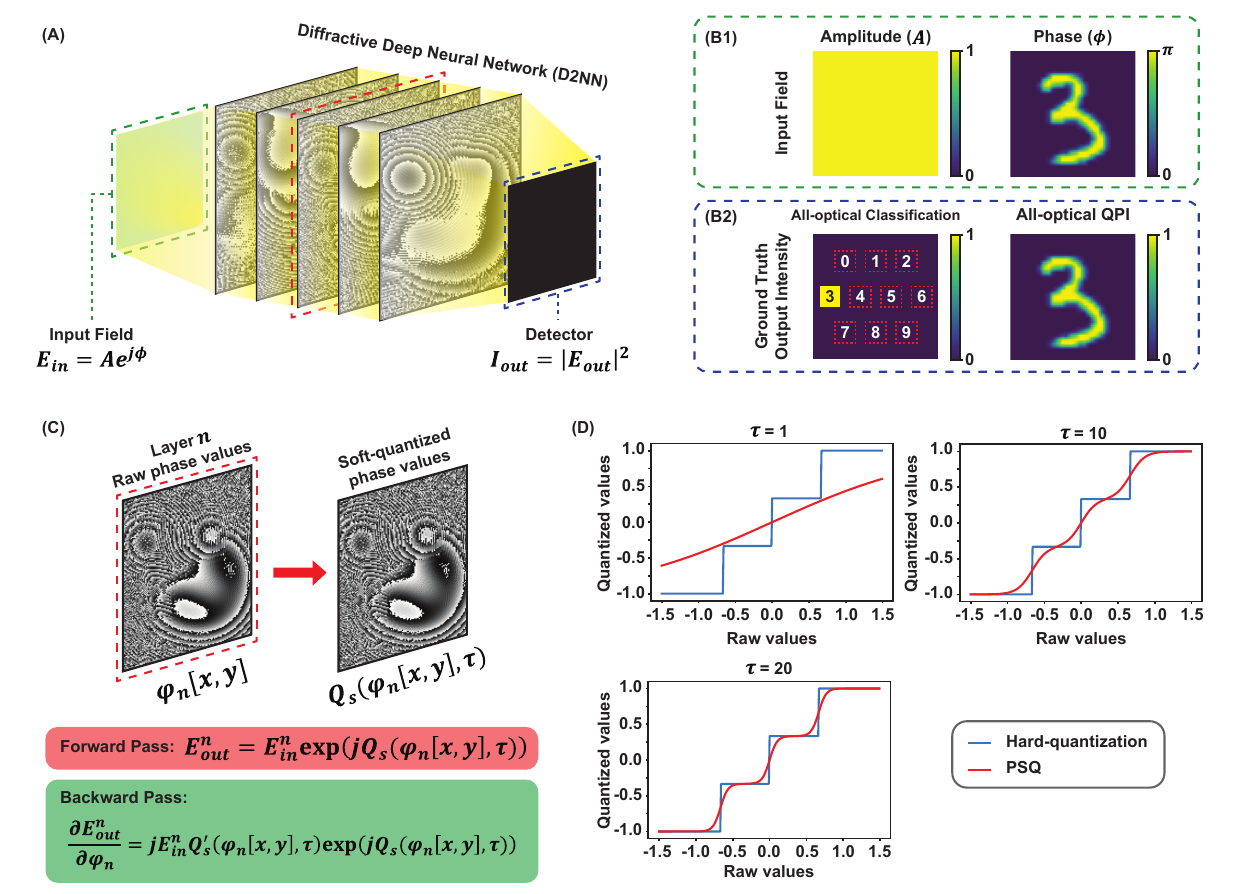}
    \caption{\textbf{Quantization-aware training of diffractive deep neural networks (D2NNs) using progressive sigmoid quantization (PSQ):} \textbf{A)} D2NN architecture: D2NNs consist of several diffractive layers. The input field passes through the layers and the detector captures the output intensity. \textbf{B1)} an example of the amplitude and phase of the input field for the MNIST dataset. The input phase contains the information of interest. \textbf{B2)} the ground truth output intensities for the two tasks considered. For the all-optical classification task, the detector region is divided into 10 patches corresponding to each class shown in red dotted lines. For the example shown, the area corresponding to digit 3 is lighted up, and the other areas have zero intensity. For the all-optical quantitative phase imaging (QPI) task, the ground truth output intensity is proportional to the input phase. \textbf{C)} the training procedure optimizing only the phase coefficients of the D2NN. During forward propagation, the raw phase weights of the $n^{\textrm{th}}$ layer $(\varphi_{n}[x,y])$ are sent through the PSQ function $Q_s(.)$. The immediate output of the layer $(E^{n}_{out})$ is obtained by modulating the input to the layer $(E^{n}_{in})$ with the soft-quantized phase coefficients as shown in the red box. During the backward propagation through the layer, the partial derivatives with respect to phase weights are computed as shown in the green box. \textbf{D)} the evolution of the PSQ function with the temperature parameter $(\tau)$. When $\tau$ increases from 1 to 20, the function gradually becomes closer to hard quantization while keeping the differentiability.}
    \label{fig:psq_with_k}
\end{figure*}

Our proposed progressive sigmoid quantization (PSQ) function is defined as,
\begin{equation}
    \label{eq:psq}
    Q_s(x,\tau) = l + \sum_{i=0}^{N-2} \Delta sig\left(\tau\left(x - l - \frac{\Delta}{2} - i \Delta\right)\right),
\end{equation}
where
\begin{equation}
    \label{eq:sigmoid}
    sig(x) = \frac{1}{1+e^{-x}},
\end{equation}
and $\tau$ is the temperature factor that changes the steepness of transition between two adjacent quantization levels (other vairables are the same as in Eq.~\ref{eq:uniform_quant}). As $\tau$ increases, the PSQ function approaches the hard-quantization function, as shown in Fig.~\ref{fig:psq_with_k}-D. When using PSQ in QuATON we start training with a smaller $\tau$ value (i.e., a relaxed quantization function). We then gradually increase $\tau$  as the training progresses, until PSQ ($Q_s$ in Eq.~\ref{eq:psq}) approaches the hard quantization function ($Q_h$ in Eq.~\ref{eq:uniform_quant}). We call this \emph{progressive training}. In this work, we consider two progressive training approaches: 1) linearly increasing temperature (PSQ-LI); and 2) treating the temperature learnable (PSQ-LT). In PSQ-LI, we start with a small $\tau$ value and we increase it linearly with the number of epochs. In PSQ-LT, we use gradient-based optimization to adjust $\tau$ as the training progresses. These two schemes are described in detail in the Methods section (under Progressive Training Schemes).

In the next section we show how QuATON can be used to train diffractive networks.

\subsection*{QuATON in Diffractive Networks}

In this section we use QuATON to train Diffractive Netowrks (D2NNs), a type of optical processors first demonstrated by Lin et al. for all optical image processing at terahertz wavelengths~\cite{Lin2018All-opticalNetworks}. D2NNs have a set of diffractive layers through which the light passes and acts as 2D fully connected networks. These diffractive layers have a set of discrete spatial locations (neurons), each having a complex transmission coefficient. The transmission coefficient of each neuron modulates the amplitude and the phase of the incoming light wave at each layer. The size of a typical D2NN neuron is in the order of half of the operating wavelength; the original terahertz D2NNs consisted of millimeter-scale neurons; D2NNs that work at visible-wavelengths --the more preferred operational range for optical imaging and image processing-- require nanometer-scale neurons. These nanometer-scale optical neurons are notoriously difficult to fabricate at full precision. The fabrication processes used to fabricate them impose constraints on the precision and the bounds of the transmission coefficients. Thus here we used QuATON to train visible-range D2NNs with quantized parameters, and thereby relaxing the required fabrication precision.

Micro-fabrication technologies, like two-photon lithography, allow manipulating the refractive index of any given 3D location of on optical substrates at diffraction limited resolution. In the context of D2NNs this capability translates to phase-only optical neurons. Thus our study considers phase only D2NNs. As shown in Fig.~\ref{fig:psq_with_k}-A let, $E_{in}[x,y]$ and $E_{out}[x,y]$, be the input and output light fields to the D2NN. At the output light field, a detector is placed to capture its intensity $I_{out}[x,y] = |E_{out}[x,y]|^2$. Let $F_{D2NN}$(.) be the function of D2NN that maps the input light field to the output light field. 
\begin{equation}
    \label{eq:d2nn_overall1}
    E_{out}[x,y] = F_{D2NN}(E_{in}(x,y)),
\end{equation}
$F_{D2NN}$(.) is the composition of individual functions two types of "layers" as shown below. 
\begin{equation}
    \label{eq:d2nn_overall2}
    F_{D2NN} = G_{out} \circ F_N \circ G \circ F_{N-1} \circ ... \circ F_n \circ G \circ ... \circ G \circ F_1 \circ G_{in},
\end{equation}
$G$'s are propagation layers, and $F$'s are modulation layers. $G_{in}$, $G$, and $G_{out}$ respectively propagates the input field to the first layer, from one layer to the next layer, and from the final layer to the detector. To model the function of $G$'s we use the Rayleigh-Sommerfeld diffraction theory (Ch. 3.5 in Goodman ~\cite{goodman2005}). Details of the propagation operation are given in Eq. (S2)-(S4) in section A of the supplementary material. Next, $F_n$, (where $n \in \{1, 2, ..., N\}$) denotes the modulation happening at the $i^{\textrm{th}}$ layer. Modulation layers, $F_n$'s consist of phase coefficients and are treated as a quantization instances for QuATON. As shown in Fig.~\ref{fig:psq_with_k}-C, during the forward propagation, the raw phase coefficients are sent through the PSQ (or DSQ) function to obtain the soft-quantized phase coefficients. Then the incoming light field to the layer is modulated with the soft-quantized phase coefficients. For the $n^{\textrm{th}}$ layer, this is given as
\begin{equation}
    \label{eq:d2nn_output}
    E^{n}_{out}[x,y] = F_{n}\left(E^{n}_{in}[x,y]\right) = E^{n}_{in}[x,y] \exp \left(j Q_s\left(\varphi_n[x,y],\tau_n\right)\right),
\end{equation}
where $E^{n}_{in}$ and $E^{n}_{out}$ are the fields immediately before and after the layer $n$. $\varphi_n$ and $\tau_n$ are the phase coefficients and quantization temperature of layer $n$ respectively. During the backpropagation, the partial derivative of the layer output with respect to phase coefficients is computed as shown in Fig.~\ref{fig:psq_with_k}-C. Similarly, for the learnable temperature case, the partial derivatives with respect to $k_n$s are also computed for the optimization of the temperature parameters. A detailed description explaining the backpropagation is also given in the supplementary materials sec. A.

In the next section, we evaluate the performance of D2NNs trained with QuATON compared to other competing methods.

\subsection*{QuATON Designs State-of-the-art Diffractive Networks Despite Quantized Weights}

As shown in Fig.~\ref{fig:psq_with_k}-A and B, we designed D2NNs for two selected physics-based tasks; \emph{all-optical classification} and \emph{all-optical quantitative phase imaging (QPI)}~\cite{mengu2022all, herath2022differentiable}. In all-optical classification, the D2NN was trained to classify phase objects whereas in all-optical QPI, it was trained to perform QPI of a phase object. For both tasks, the information of interest is in the phase of the input field. Therefore, the input light-field to the D2NN models were constructed by placing images (from the datasets) in the phase after scaling into the range $[0, \pi]$. The amplitude of the input field was set to one throughout the field. An example of the input field to a D2NN is shown in Fig.~\ref{fig:psq_with_k}-B1 and the field can be given as,
\begin{equation}
    \label{eq:d2nn_input}
    E_{in}[x,y] = e^{j \phi[x,y]},
\end{equation}
where $\phi[x,y]$ is the input phase, which is an image from the dataset scaled to $[0,\pi]$. The D2NN training details and the two physics-based tasks are explained in detail in the Methods section.

For each task we experimented on multiple datasets. Our experiments were organized as follows. For each task and each dataset, we first trained D2NNs with full precision weights without quantization. These models (denoted by FP) established heuristic upper limits for the performance of a specific task on a particular dataset. We then hard quantized the weights of the pre-trained full precision models, generating results for post-quantized (PQ) weights. We then trained our D2NNs using two existing QAT methods used to train optical neurons, straight-through estimator (STE)~\cite{bengio2013estimating}, and Gumbel-Softmax based quantization (GS)~\cite{Li2022Physics-AwareNetworks}. These experiments, i.e., PQ, STE and GS, set the baseline of the current state-of-the-art (SOTA). A detailed description of these methods is given in the Methods section. Finally we trained D2NNs using QuATON. We experimented four QuATON variations. We first used the same soft quantization and training mechanisms in Differentiable soft quantization (DSQ)~\cite{Gong2019DifferentiableNetworks} (note that DSQ has not been used to train optical neurons before). We then used our proposed PSQ while keeping the temperature ($\tau$) fixed (denoted by PSQ-FT). Last we used PSQ with our two progressive training approaches, i.e. with a linearly increasing temperature (denoted by PSQ-LI) and the learnable temperature (denoted by PSQ-LT). 

In following subsections we present the results of these experiments for the two physics-based tasks, all-optical classification, and all-optical QPI.


\begin{figure*}
        \centering
        \includegraphics[width=\linewidth]{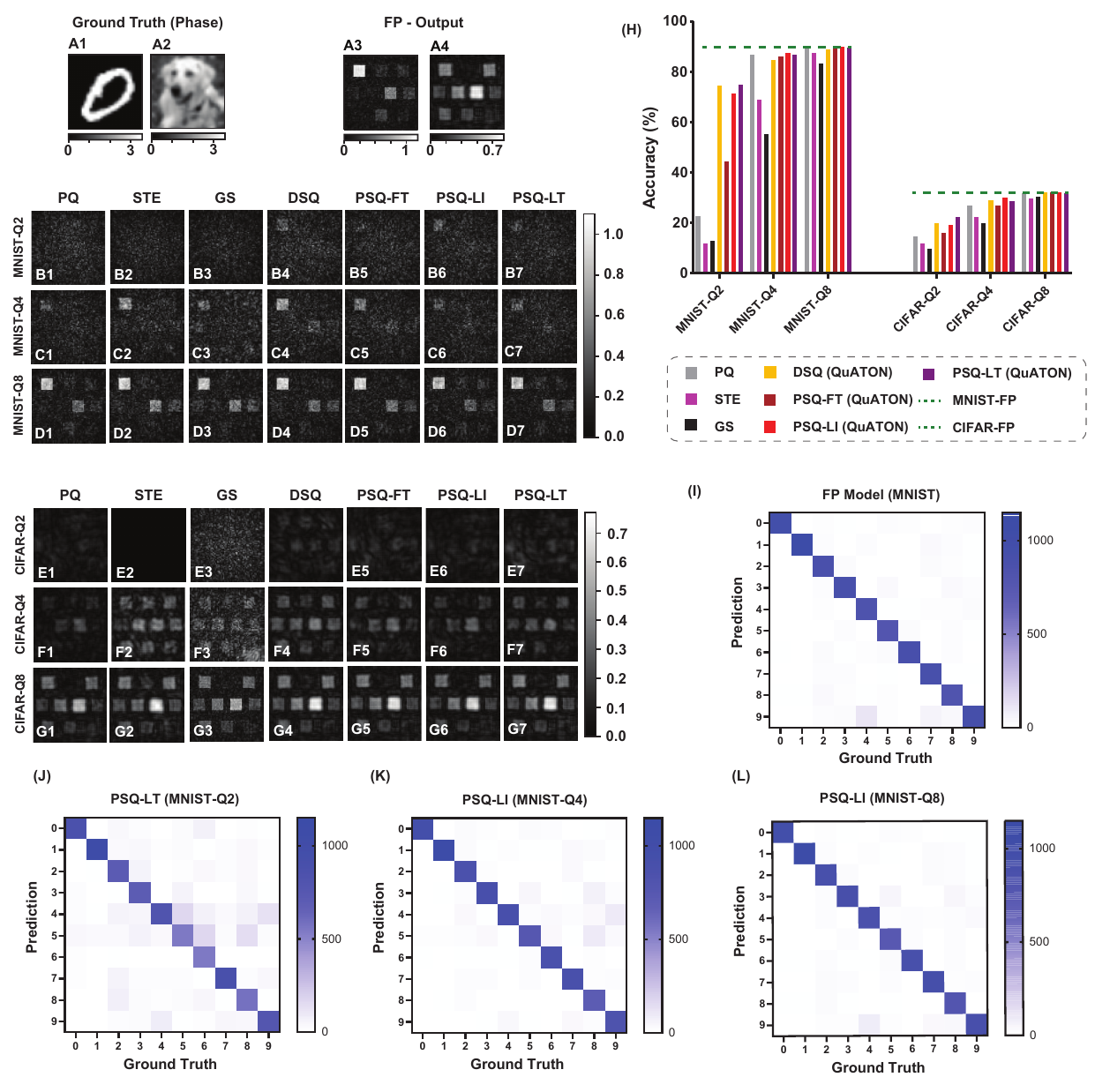}
        \caption{\textbf{All-optical classification results:} \textbf{A1-A2)} two examples of the phase of the incoming wave to the D2NN for the two datasets considered. \textbf{A3-A4)} output intensities for the D2NNs trained with full precision (FP) weights. \textbf{B-G)} classification results for the quantization-aware trained D2NNs for each of the examples. Each row named as \textbf{\textit{x}-Q\textit{n}} shows the results for dataset \textbf{\textit{x}} $\in \{\textrm{MNIST, CIFAR10}\}$, using D2NNs trained with \textbf{\textit{n}}-level quantized weights ($\textbf{\textit{n}} \in \{2, 4, 8\}$). Each column corresponds to different QAT methods considered which are stated above row B).  \textbf{H)} comparison of quantitative results (classification accuracy over the test set of each dataset). \textbf{I-L)} confusion matrices for the FP model and the best-performing methods for each quantization level for the MNIST dataset.}
        \label{fig:qpi_classifier_results}
    \end{figure*}

\subsubsection*{All-optical Classification}

We experimented on two classification datasets, MNIST and CIFAR10 as shown in Table \ref{tab:d2nn-classifier}. D2NN weights were quantized with 2, 4, and 8 quantization levels (Q2, Q4, and Q8). The full precision D2NN achieved $\approx$ 90\% accuracy for the MNIST data set. However, for the more challenging CIFAR10 dataset, the classification accuracy was only 31.88\%, even for the full precision model. We observe that 8 quantization levels were sufficient to achieve full precision accuracy in both datasets. Therefore, for 8 quantization levels, even without quantization-aware training frameworks, we achieved performance as good as the full precision model. However, for 4 and 2 levels of quantization, our QuATON variants (DSQ, PSQ-FT, PSQ-LI, and PSQ-LT) showed a clear improvement in performance. PSQ variants outperformed DSQ by a small margin on all datasets and all quantization levels. The qualitative results are shown in Figure \ref{fig:qpi_classifier_results}. Of note, for 2 quantization levels in the MNIST dataset, QuATON methods showed more than 50\% increase in accuracy compared to current SOTA. 

\begin{table}
\centering
\caption{Quantitative results of all-optical classification using D2NNs (classification accuracy).}
\label{tab:d2nn-classifier}
\begin{tabular}{@{}lcclccclccclccc@{}}
\toprule
\multirow{2}{*}{\textbf{Method}} & \multicolumn{1}{c}{\multirow{2}{*}{\textbf{\begin{tabular}[c]{@{}c@{}}Proposed\\ by\end{tabular}}}} & \multicolumn{1}{c}{\multirow{2}{*}{\textbf{\begin{tabular}[c]{@{}c@{}}Used for\\ Optical Neurons in\end{tabular}}}} & \multirow{2}{*}{}  & \multicolumn{3}{c}{\textbf{MNIST}}                     & \textbf{} & \multicolumn{3}{c}{\textbf{CIFAR10}}                   \\ \cmidrule(l){5-11} 
                            & \multicolumn{1}{c}{}                                                                                & \multicolumn{1}{c}{}                         & \textbf{} & \textbf{Q2}      & \textbf{Q4}      & \textbf{Q8}      & \textbf{} & \textbf{Q2}      & \textbf{Q4}      & \textbf{Q8}      \\ \midrule
FP                             &     -      &  -  &          &\multicolumn{3}{c}{89.99\%}                            &           & \multicolumn{3}{c}{31.88\%}                            \\ \midrule
PQ              & - &       \cite{mengu2022all}          &           & 21.84\%          & 86.89\%          & 90.06\%          &           & 14.54\%          & 26.79\%          & 31.89\%          \\
STE               & \cite{bengio2013estimating} &      \cite{gu2020roq}         &           & 11.98\%          & 69.05\%          & 87.43\%          &           & 11.97\%          & 22.55\%          & 29.80\%          \\
GS              & \cite{jang2017categorical} &        \cite{Li2022Physics-AwareNetworks}         &           & 12.79\%          & 55.19\%          & 83.50\%          &           & 9.86\%           & 19.88\%          & 30.62\%          \\ \midrule
DSQ              & \cite{Gong2019DifferentiableNetworks} &        This paper        &           & 74.75\%          & 84.69\%          & 89.13\%          &           & 19.98\%          & 29.07\%          & 32.10\%          \\
PSQ-FT    & This paper &        This paper        &           & 44.53\%          & 86.34\%          & 89.62\%          &           & 16.14\%          & 26.91\%          & 32.08\%          \\
PSQ-LI     & This paper &        This paper       &           & 71.31\%          & \textbf{87.73\%} & \textbf{90.08\%} &           & 19.23\%          & \textbf{30.18\%} & \textbf{32.35\%} \\
PSQ-LT     & This paper &       This paper       &           & \textbf{75.03\%} & 87.06\%          & 89.76\%          &           & \textbf{22.27\%} & 28.83\%          & 32.01\%          \\ \bottomrule
\end{tabular}
\end{table}

\subsubsection*{All-optical Quantitative Phase Imaging (QPI)}

\begin{figure*}
        \centering
        \includegraphics[width=\linewidth]{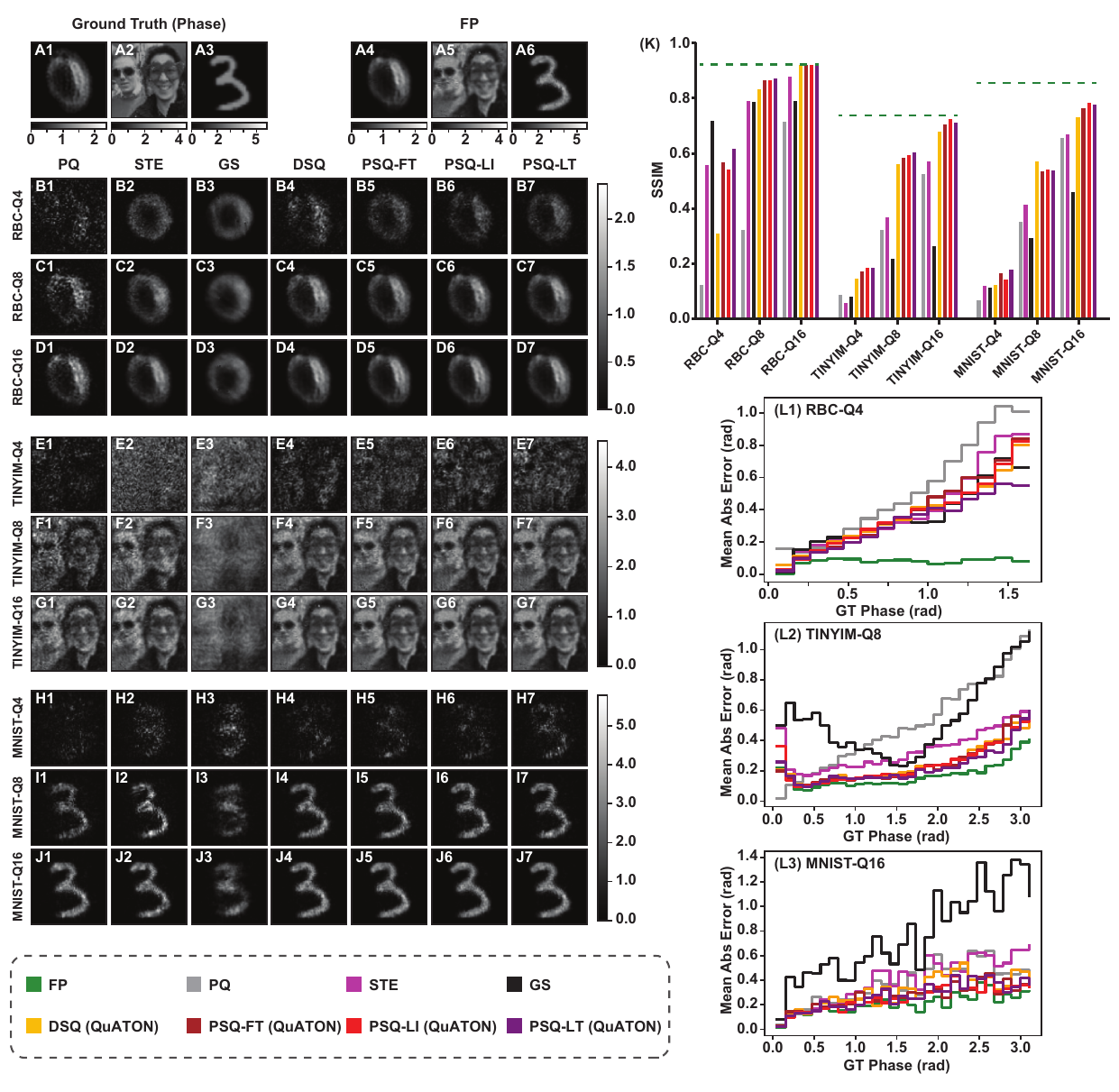}
        \caption{\textbf{All-optical quantitative phase imaging results:} \textbf{A1)-A3)} three examples of the phase of the incoming wave to the D2NN for the three datasets considered. \textbf{A4-A6)} output intensities $\times \pi$ for the D2NNs trained with full precision (FP) weights. \textbf{B)-J)} QPI results for quantization-aware trained D2NNs for each of the examples. Each row named as \textbf{\textit{x}-Q\textit{n}} shows the results for dataset \textbf{\textit{x}} $\in \{\textrm{RBC, TINYIM, MNIST}\}$, using D2NNs trained with \textbf{\textit{n}}-level quantized weights ($\textbf{\textit{n}} \in \{4, 8, 16\}$). Each column corresponds to different QAT methods considered which are stated above row B). Note that all results are given as $\textrm{output intensity} \times \pi$. \textbf{K)} the comparison of the quantitative results (mean SSIM over the test set of each dataset). \textbf{L1)-L3)} mean absolute phase error variation against ground truth phase for \textbf{RBC-Q4, TINYIM-Q8,} and \textbf{MNIST-Q16} cases respectively. These plots are shown for the given examples in the figure.}
        \label{fig:qpi_results}
    \end{figure*}

All-optical quantitative phase imaging translates the phase information of the input light field to the intensity at the output light field. It is thus an image translation task that's more challenging than the previous classification task. As shown in Table \ref{tab:d2nn-qpi} we experimented on three datasets, MNIST, TinyImageNet, and RBC (red blood cells). D2NN weights were quantized with 4, 8, and 16 quantization levels (Q4, Q8, and Q16).  The comparison of the quantitative results (using the structural similarity index measur - SSIM~\cite{wang2004image}) for this task is given in Table~\ref{tab:d2nn-qpi}. The full precision D2NN (FP) achieved an SSIM of 0.8560, 0.7385, and 0.9227 on the MNIST, TinyImageNet, and RBC datasets respectively. Post quantization of FP models severely degraded the model performances especially for Q4 and Q8 quantization levels (see row PQ in Table \ref{tab:d2nn-qpi}).  STE and GS behaved differently on each dataset. On the MNIST and RBC datasets, STE improved the SSIM while for the TinyImageNet dataset the improvement was not consistent. For the MNIST and TinyImageNet datasets, GS in fact degraded the SSIM values (compared to the PQ baseline), but surprisingly improved by a large margin for the RBC dataset. Our QuATON variants (DSQ, PSQ-FT, PSQ-LI, and PSQ-LT) clearly and consistently improved the SSIM on all datasets and quantization levels (compared to the PQ baseline). In almost all cases, PSQ variants with progressive training (i.e., PSQ-LI and PSQ-LT) outperformed the others. Except for the RBC-Q4 case, at least one QuATON variant over performed the current SOTA (i.e., PQ, STE and GS). Surprisingly, GS showed the highest SSIM for the RBC-Q4 case. For a deeper analysis of these results, we consider the qualitative results shown in Fig.~\ref{fig:qpi_results}.

Considering the qualitative results for the RBC dataset with 4-level quantization (RBC-Q4), although GS (Fig.~\ref{fig:qpi_results}-B3) resulted a higher SSIM value, it had failed to capture the morphological features of the red blood cell as well as the PSQ methods (Figs.~\ref{fig:qpi_results}-B6 and B7). Further analysis revealed that for the RBC dataset, the D2NNs trained with GS results mode collapse. Additional examples showing this are given in supplementary materials Fig. S1. This is also evident from the mean phase error plot shown in Fig.~\ref{fig:qpi_results}-L1. It can be clearly seen that PSQ-LT has a lower mean phase error throughout the entire phase range, especially for higher phase values compared to other methods for the considered example. For RBC-Q8 and RBC-Q16 (Fig.~\ref{fig:qpi_results}-C and D), our methods perform similar to the FP model.

For TinyImageNet with 4-level quantization (TINYIM-Q4), none of the methods produces good results. However, it can be noted that our QuATON variants (Figs.~\ref{fig:qpi_results}-E4-E7) have managed to capture some of the features of the input phase. For the TINYIM-Q8 and TINYIM-Q16 cases, QuATON variants  showed better performance than the current SOTA appraoches. The qualitative results for the MNIST dataset also show similar trends as the TinyImageNet. However, for MNIST-Q4, PSQ has managed to capture the shape of the digit much better compared to other methods (Fig.~\ref{fig:qpi_results}-H). Fig.~\ref{fig:qpi_results}-L3 shows the phase error plot for MNIST-Q16. This shows that similar to TinyImageNet, although PSQ and DSQ perform on par with each other, for certain phase ranges (E.g. from $1.5$ rad to $2.5$ rad) PSQ methods have lower errors than DSQ. For further validity of these observations, additional examples for each dataset are shown in the supplementary materials Fig. S2. Furthermore, the progressive quantization of D2NN phase coefficients for the TINYIM-Q8 case using the PSQ-LT method is shown in supplementary materials Fig. S3.

\begin{table}[t]
    \centering
    \caption{Quantitative results of all-optical quantitative phase imaging using D2NNs. The SSIM values are given for FP models and quantized models trained using different QAT methods.}
    \label{tab:d2nn-qpi}
    \begin{tabular}{@{}llccclccclccc@{}}
    \toprule
    \multirow{2}{*}{\textbf{Method}} & \textbf{} & \multicolumn{3}{c}{\textbf{MNIST}}                  & \textbf{} & \multicolumn{3}{c}{\textbf{TinyImageNet}}           &  & \multicolumn{3}{c}{\textbf{RBC}}                    \\ \cmidrule(l){2-13} 
                                     & \textbf{} & \textbf{Q4}     & \textbf{Q8}     & \textbf{Q16}    & \textbf{} & \textbf{Q4}     & \textbf{Q8}     & \textbf{Q16}    &  & \textbf{Q4}     & \textbf{Q8}     & \textbf{Q16}    \\ \midrule
    FP                               &           & \multicolumn{3}{c}{0.8560}                          &           & \multicolumn{3}{c}{0.7385}                          &  & \multicolumn{3}{c}{0.9227}                          \\ \midrule
    PQ                               &           & 0.0674          & 0.3526          & 0.6555          &           & 0.0870          & 0.3234          & 0.5258          &  & 0.1232          & 0.3214          & 0.7140          \\
    STE                              &           & 0.1200          & 0.4138          & 0.6698          &           & 0.0570          & 0.3689          & 0.5721          &  & 0.5578          & 0.7898          & 0.8774          \\
    GS                               &           & 0.1114          & 0.2938          & 0.4595          &           & 0.0793          & 0.2182          & 0.2610          &  & \textbf{0.7188} & 0.7854          & 0.7891          \\ \midrule
    DSQ (QuATON)                             &           & 0.1207          & \textbf{0.5701} & 0.7321          &           & 0.1433          & 0.5610          & 0.6782          &  & 0.3095          & 0.8333          & \textbf{0.9208} \\
    PSQ-FT (QuATON)                    &           & 0.1653          & 0.5348          & 0.7627          &           & 0.1709          & 0.5845          & 0.7038          &  & 0.5678          & 0.8646          & 0.9199          \\
    PSQ-LI (QuATON)                    &           & 0.1411          & 0.5412          & \textbf{0.7822} &           & \textbf{0.1854} & 0.5924          & \textbf{0.7237} &  & 0.5394          & 0.8663          & 0.9206          \\
    PSQ-LT (QuATON)                    &           & \textbf{0.1772} & 0.5374          & 0.7759          &           & 0.1832          & \textbf{0.6042} & 0.7125          &  & 0.6156          & \textbf{0.8723} & 0.9202          \\ \bottomrule
    \end{tabular}
\end{table}
\section*{Summary}
\label{sec:discussion}

In this study, we presented a quantization-aware training (QAT) framework for optical neutrons called QuATON. QuATON is based on a soft differentiable quantization function and a progressive training approach. We demonstrated the results of this method on two physics-based tasks performed by diffractive networks (D2NNs).

Our results show that QuATON outperforms the post-quantization, straight-through estimator, and Gumbel-Softmax which are the current state-of-the-art(SOTA) QAT methods used for optical neutrons. D2NNs with quantized weights trained using our method managed to achieve similar performance to a model with full-precision, using a smaller number of quantization levels (4 levels for all-optical classification and 16 levels for all-optical QPI). Furthermore, progressive training based methods outperformed fixed temperature based methods in almost all cases. This shows that progressive training is significant when the number of quantization levels is smaller. Altogether, this comprehensive evaluation of QAT methods for optical neural architectures sheds light on which method to choose to achieve a desired level of performance for a task while being constrained to a given precision of physical parameters.

Although not demonstrated in this work, our method can easily be extended to non-uniform quantization by combining multiple PSQ functions. It is important to note that this study presents a numerical simulation of the performance of the D2NNs. The physical realization process of these networks can introduce other noise and artifacts that should be considered during the training process. In future work, we aim to include other types of noise in the training process and fabricate them to experimentally validate the results.

In conclusion, our QAT framework addresses the issue of lack of precision in fabrication methods, optical devices, and analog-to-digital/ digital-to-analog conversions. We believe that this work lays the foundation upon which optical neurons can be physically realized for challenging vision applications in the future.
\section*{Materials and Methods}
\label{sec:methods}

\subsection*{Progressive Training Schemes}
\label{subsec:progressive_training}

This section describes the two progressive training schemes used in QuATON: 1) linearly increasing temperature (PSQ-LI) and 2) learnable temperature (PSQ-LT).

\subsubsection*{Linearly Increasing Temperature}
In this approach, we start by setting $\tau$ to a small value and increase it linearly over the training epochs. The scheduling of $\tau$ has three hyperparameters; \emph{the initial value of $\tau$} $(\tau_0)$, \emph{the increment step size} $(\Delta \tau)$, and \emph{the interval between two consecutive increments} $(\Delta t)$. Using these hyperparameters, the temperature at epoch $t$ $(\tau(t))$ can be given as
\begin{equation}
    \tau(t) = \tau_0 + \left\lfloor \frac{\Delta \tau}{\Delta t}  \right\rfloor t.
\end{equation}

\subsubsection*{Learnable Temperature}
In learnable temperature scheduling, we optimize $\tau$ through backpropagation. In this context, we define \emph{quantization instances} (QIs), each having a separate PSQ function that is characterized by its own specific temperature factor. A QI can consist of either a single trainable parameter or a set of such parameters. Consider an example of an ONA with $M$ number of QIs. For the $m^{\textrm{th}}$ quantization instance, the temperature $(\tau_m)$ is defined as
\begin{equation}
    \label{eq:km}
    \tau_m = \frac{1}{|k_m|+\gamma},
\end{equation}
where $k_m$ is a trainable parameter and $\gamma$ is a constant setting the upper bound of $\tau_m$. To promote the progressive nature in the optimization process of $\tau_m$, we introduce a regularization term in the loss computation. For the $t^{\textrm{th}}$ training epoch, the regularization term $R_t$ is given by
\begin{equation}
    R_t(\textbf{k}) = \lambda_1 s_t \left(\lVert \textbf{k} \rVert^2_2 - \lambda_{2}^{2}\right).
\end{equation}
In this equation, $\textbf{k} = [k_1, k_2, \ldots, k_M]^T$ is the vector containing the $k_m$ parameters (Eq. \ref{eq:km}) of each QI in the model and $\lambda_1, \lambda_2$ are hyperparameters. $s_t$ is a constant that is updated every $\beta$ epoch as $s_t = 2^{\lfloor t/\beta \rfloor}$. This term forces the temperature to increase every $\beta$ epoch while allowing it to be updated through backpropagation. The overall loss is computed as


\begin{equation}
    \mathcal{L}(y,\hat{y};\boldsymbol{\Theta},\textbf{k}) = \mathcal{L}_f(y,\hat{y};\boldsymbol{\Theta},\textbf{k}) + R_t(\textbf{k})
    \label{eq:overall_loss}
\end{equation}
where $y$ is the ground truth, $\hat{y}$ is the predicted output, $\mathcal{L}_f$ is the loss function specific to the task, and $\boldsymbol{\Theta}$ is the set of trainable parameters of the ONA. Note that the regularization term ($R_t(\textbf{k})$) is added to the loss only when the temperature is learnable.

\subsection*{All-Optical Classification}
We consider the all-optical classification of phase images using a D2NN. The task is evaluated on two datasets: MNIST digits~\cite{deng2012mnist} and CIFAR10~\cite{cifar10}. Since both datasets have 10 classes, in this task, the output plane has 10 spatially separated patches, as shown in Fig.~\ref{fig:psq_with_k}-B2. The patch with the highest mean intensity determines the class. During training, the objective is to concentrate most of the light on the patch corresponding to the ground truth class and suppress the light that scatters to other patches.  The loss function for this task is given in Eq. (\ref{eq:classification_loss}),

\begin{equation}
    \label{eq:classification_loss}
\mathcal{L}_f(Y, |E_{out}|^2; \varphi, \textbf{k}) = \frac{1}{n} \sum_{i=1}^{n} \left( \text{SE}_i \times \left( 1 - \frac{Y_i}{11} \right) \right)
\end{equation}

where $\text{SE}_i = \left( Y_i - |E_{out_i}|^2 \right)^2$ is the squared error (SE) for the \( i \)-th pixel. $Y$ is the label map corresponding to the ground truth digit. For example, as shown in Fig. \ref{fig:psq_with_k},  only the patch corresponding to digit 3 will equal 1 in the label map (Y) for digit 3. In essence, the loss function is a weighted mean squared error, where more weight is given to the non-target region to penalize light scattering to those areas.

\subsection*{All-Optical Quantitative Phase Imaging}
In the all-optical QPI task, we consider three datasets; MNIST digits~\cite{deng2012mnist}, TinyImageNet~\cite{Le2015TinyIV}, and an experimentally collected red blood cell (RBC) dataset. In this task, the D2NN is trained with the objective
\begin{equation}
    \label{eq:qpi_loss}
    \min_{\boldsymbol{\varphi},\mathbf{k}} \mathcal{L}\left(|E_{out}[x,y]|^2, \frac{\phi(x,y)}{\pi};\boldsymbol{\varphi}, \textbf{k} \right) 
\end{equation}
such that the output intensity is proportional to the input phase. Here, $\boldsymbol{\varphi}$ denotes the phase coefficients of the D2NN, $\mathcal{L}$ is the overall loss, $E_{out}[x,y]$ is the output field of the D2NN, and $\textbf{k}$ is the vector containing $k$ parameters of each D2NN layer. The overall loss is computed according to Eq.~\eqref{eq:overall_loss} and reverse Huber loss~\cite{Zwald2012TheEffect} is used as the loss function $\mathcal{L}_f$.

\subsection*{Training Details}

In both tasks, only the phase coefficients of the D2NNs are optimized. During the training process, we apply PSQ for QAT of the phase coefficients. For a given task and a given dataset, we first train a D2NN with full-precision (FP) weights without quantization for 100 epochs (200 epochs for CIFAR10 dataset). Then, starting from FP weights, we apply PSQ to train the D2NN with quantized phase coefficients for another 100 epochs. Since we consider only the phase coefficients, they lie in the range $[0, 2\pi)$ rad. However, the trained phase coefficients of the FP model are unwrapped phase values, thus distributing them over several phase cycles. As a better initialization for the QAT process, we wrap the FP weights to bring them to the range $[0, 2\pi)$. This prevents weights outside of $[0, 2\pi)$ from being clamped to the lowest or highest quantization level at the beginning of the QAT process.

Furthermore, during QAT we limit the range of quantization levels to $[0, 1.99\pi]$ rad since $0$ rad and $2\pi$ rad correspond to the same phase shift. For example, if we consider phase weights with 4-level quantization, the corresponding quantization levels will be $[0, 0.663\pi, 1.327\pi, 1.990\pi]$ rad. However in the all-optical classification task, for 2-level quantization, we use the levels $[0, \pi]$. We evaluate three versions of PSQ and compare their performance for both tasks: 1) PSQ with keeping $\tau$ fixed (PSQ-FT), 2) PSQ with linearly increasing $\tau$ (PSQ-LI), and 3) PSQ with learnable $\tau$ (PSQ-LT). 

For the all-optical classification task, we use 7-layer D2NNs trained with 2, 4, and 8 quantization levels for each case. For the all-optical QPI task, we train each D2NN with three different quantization levels (4, 8, and 16). For this task, we use 5-layer D2NNs for both TinyImageNet and RBC datasets, and 7-layer D2NNs for the MNIST dataset. Further details on D2NN specifications and the training process are included in the supplementary materials sec. B.

\subsection*{Comparison of Performance}

We compare the performance of PSQ for the considered tasks with several QAT methods, the FP model, and the post-quantized FP model. For a fair comparison, we initialize the D2NN phase weights with FP weights before QAT using all the methods.

\subsubsection*{Post-Quantization (PQ)} In this, the phase weights of the FP model are directly quantized using Eq.~\eqref{eq:uniform_quant}.

\subsubsection*{Straight-through Estimator (STE)~\cite{bengio2013estimating}} In this method, Eq.~\eqref{eq:uniform_quant} is used in the forward pass while during the backpropagation, gradients are computed as,
\begin{equation}
    \label{eq:ste_der}
    \frac{\partial Q_h}{\partial x} = 1 \quad \forall x.
\end{equation}
where $x$ is any trainable parameter in the network.

\subsubsection*{Gumbel-Softmax based Quantization (GS)} This method uses Gumbel-Softmax as a soft quantization function, and this has been demonstrated as a QAT technique for D2NNs by Li et al.~\cite{Li2022Physics-AwareNetworks}. When training D2NNs using GS, we use the linear annealed temperature schedule (temperature starts from 50 and decreases by 0.5 each epoch), which they have mentioned in the paper for best-performing results.


\subsubsection*{Differentiable Soft Quantization (DSQ)} This method was proposed by Gong et al.~\cite{Gong2019DifferentiableNetworks} as a QAT method for DNNs. In their original work, DSQ is implemented with a learnable temperature and lower and upper bounds of the quantization range. However, since we consider the phase weights of D2NNs in the range $[0, 2\pi)$, we keep the lower and upper bounds of the quantization range fixed. 

\bibliography{sample}



\section*{Acknowledgements}

This work was supported by the Center for Advanced Imaging at Harvard University (H.K., R.H., and D.N.W.), and NIH R21-MH130067-01 (D.N.W.). D.N.W. further acknowledge support from the John Harvard Distinguished Science Fellowship Program within the FAS Division of Science of Harvard University. Q.Y., P.T.S., and E.S.B. acknowledge support from Fujikura Inc. A.M. and P.T.S. also acknowledge support from NIH R01HL158102. E.S.B. further acknowledges the support from HHMI, John Doerr, and Lisa Yang.


\section*{Author contributions statement}

H.K. and R.H. developed the methods and conducted the numerical experiments; H.K., R.H., and D.N.W. analysed results; Q.Y., T.N.,  Hi.K. and Y.K. formulated the fabrication requirements for optical neurons; A.M. prepared the RBC dataset; P.T.S. advised Q.Y., T.N. and A.M.; Y.K. advised T.N. and Hi.K.; E.S.B. advised Q.Y.; D.N.W. advised H.K. and R.H. and supervised the project.

\section*{Additional information}
 
\textbf{Competing interests:} The authors declare that they have no competing interests. 


\clearpage

\appendix

\newpage

\renewcommand\thefigure{\arabic{figure}}
\setcounter{figure}{0}
\setcounter{equation}{0}

\renewcommand{\thefigure}{S\arabic{figure}}
\renewcommand{\thetable}{S\arabic{table}}
\renewcommand{\theequation}{S\arabic{equation}}

\begin{center}%
    \Large
    \textbf{Progressive Sigmoid Quantization Framework for Optical Neural Architectures} \\
    \vspace{0.5em}Supplementary Material \\
    \vspace{1.0em}
\end{center}

\appendix

\section{Simulating Light Propagation}
\setcounter{page}{1}
\label{sec:light_prop}

This section gives a detailed information on light propagation through the D2NNs and backpropagation to optimize the D2NN and PSQ paratmeters. During the QAT, the forward propagation through the immediate output of the $n^{\textrm{th}}$ layer is given by Eq.~\eqref{eq:d2nn_output}, which is given below.
\begin{equation}
    \label{eq:d2nn_output2}
    E^{n}_{out}[x,y] = F_{n}\left(E^{n}_{in}[x,y]\right) = E^{n}_{in}[x,y] \exp \left(j Q_s\left(\varphi_n[x,y],\tau_n\right)\right)
\end{equation}
Here, $E^{n}_{out}[x,y]$ is the modulated output field from the $n^{\textrm{th}}$ layer of the D2NN, $E^{n}_{in}[x,y]$ is the input field to the $n^{\textrm{th}}$ layer. $[x,y]$ discrete coordinates correspond to the location of a neuron on the D2NN layer. $\varphi[x,y]$ is the phase coefficient of a neuron, $Q_s(.)$ is the PSQ function, and $\tau_n$ is the temperature factor of layer $n$. For propagating this field to the next layer, we use the Rayleigh-Sommerfeld diffraction formulation~\cite[ch. 3.5]{goodman2005}. Using this, the input field to neuron at $[x_i,y_i]$ of the $(n+1)^{\textrm{th}}$ layer is given by
\begin{equation}
    \label{eq:rs_prop}
    E^{n+1}_{in}[x_i,y_i] = \sum_{x,y} E^{n}_{out}[x,y] \Delta A^{n}[x,y] w^{n}_{x,y,i,z},
\end{equation}
where $\Delta A^{n}[x,y]$ is the area of the neuron at $[x,y]$ and 
\begin{equation}
    \label{eq:prop_tf}
    w^{n}_{x,y,i,z} = \left(\frac{z}{r^{2}}\right) \left(\frac{1}{2\pi r} + \frac{1}{j\lambda}\right) \exp \left(j \frac{2\pi r}{\lambda}\right).
\end{equation}
In this, $z$ is the distance between the two layers, $\lambda$ is the wavelength, and $r=r_{x,y,i,z}=\sqrt{(x_i-x)^2 + (y_i-y)^2 + z^2}$. For the simplicity, we write the propagation in Eq.~\eqref{eq:rs_prop} as
\begin{equation}
    \label{eq:prop_simplified}
    E^{n+1}_{in}[x_i,y_i] = G\left(E^{n}_{out}[x,y],x_i,y_i,z\right).
\end{equation}
For the special cases $n=0$ (i.e. propagation of input field to the first layer), and $n=N$ (i.e. propagation from the last layer to the detector) Eq.~\eqref{eq:prop_simplified} becomes,
\begin{equation}
    \begin{aligned}
        E^{1}_{in}[x_i,y_i] &= G_{in}\left(E_{in}[x,y],x_i,y_i,z_{in}\right) \\
        E_{out}[x_i,y_i] &= G_{out}\left(E^{N}_{out}[x,y],x_i,y_i,z_{out}\right),
    \end{aligned}
\end{equation}
Here $z_{in}$ and $z_{out}$ are the distance between the input plane and the first layer, and the distance between the last layer and the detector. In the simulation, we use the angular spectrum method~\cite{Ratcliffe1956, goodman2005} for an efficient implementation of this propagation.

For the optimization process, we need to compute the gradients of the field, with respect to the D2NN parameters and $k_n$ parameters of each layer (used to compute the temperature of each layer as described in the Methods and Materials section). These gradients can be computed using the chain rule as following.
\begin{equation}
    \begin{aligned}
        \label{eq:chain_rule}
        \frac{\partial E^{n+1}_{in}}{\partial \varphi_{n}} &= j G'\left(E^{n}_{out},z\right) E^{n}_{in} e^{jQ_s\left(\varphi_n, \tau_n\right)} \partial_{\varphi_n}Q_s\left(\varphi_n, \tau_n\right)\\
        \frac{\partial E^{n+1}_{in}}{\partial k_{n}} &= j G'\left(E^{n}_{out},z\right) E^{n}_{in} e^{jQ_s\left(\varphi_n, \tau_n\right)} \partial_{\tau_n}Q_s\left(\varphi_n, \tau_n\right) \partial_{k_n} \tau_n
    \end{aligned}
\end{equation}

The partial derivatives terms in the above equations are computed as
\begin{equation}
    \begin{aligned}
        \label{eq:pd_psq}
        \frac{\partial Q_s}{\partial \varphi_n} &= \Delta \tau_n \sum_{i=0}^{N-2} sig'\left(\tau_n \left(\varphi_n - \beta_i\right)\right) \\
        \frac{\partial Q_s}{\partial \tau_n} &= \Delta \sum_{i=0}^{N-2} \left(\varphi_n - \beta_i\right) sig'\left(\tau_n \left(\varphi_n - \beta_i\right)\right) \\
        \frac{\partial \tau_n}{\partial k_n} &= \frac{-1}{\left(|k_n|+ \gamma\right)^2} \frac{d|k_n|}{dk_n}, \\
    \end{aligned}
\end{equation}
where $\beta_i = l + (i+0.5) \Delta$ and $sig'(.)$ is the derivative of the sigmoid function given by
\begin{equation}
    \label{eq:sig_der}
    sig'(x) = sig(x) \left(1 - sig(x)\right).
\end{equation}
All the other symbols have the same meanings as in the main text. Note that the partial derivatives with respect to $k_n$ are computed only for the PSQ-LT method. We implement the entire training pipeline using pytorch automatic differentiation~\cite{paszke2017automatic}.

\section{Training Details}
\label{sec:train_details}

This section gives additional details regarding the training process and specifications of D2NNs considered in the study. 

\subsection{D2NN Specifications}

For both the tasks, D2NNs operating in the wavelength $\lambda = 632.8$ nm with neuron size of $0.5\lambda \times 0.5\lambda$ are used. For the MNIST digit dataset (for both tasks), 7-layer D2NNs are used with $64 \times 64$ neurons per layer. Input field of view (FoV) is the same as a layer size. The distance between two adjacent layers and the distance between the input FoV and the first layer are set to $5.3\lambda$, where the distance between the final layer and the detector is set to $9.3\lambda$.

For the tinyimagenet and red blood cell datasets (for all-optical QPI), 5-layer D2NNs with $200 \times 200$ neurons per layer are used. The input FoV in these cases are set to $2.5$ times smaller than D2NN layers, and the spacing between the layers, between the input FoV and the first layer, and between the last layer and the detector are all set to $40\lambda$. These distances are same for the CIFAR10 dataset (all-optical classification). However, D2NNs with 7 layers, each having $64 \times 64$ neurons and same size as the input FoV are used.

\subsection{Training and Performance Evaluation}

All the models are trained using the Adam optimizer~\cite{kingma2014adam}. During the training process, PSQ is used to quantize the D2NN parameters. However, during validation and testing phases, the trained parameters are quantized using the hard-quantization function.

Each dataset considered is seperated into three partitions; train, validation, and test sets. The models are trained on the train set for 100 epochs (200 epochs for CIFAR10) using full-precision parameters. Starting from this model, QAT is performed for another 100 epochs. Then the trained model of the epoch with the highest performance for the validation set is chosen and evaluated on the test set. The same method is used to evaluate the performance for other QAT methods considered. 

\begin{figure*}
    \centering
    \includegraphics[width=\linewidth]{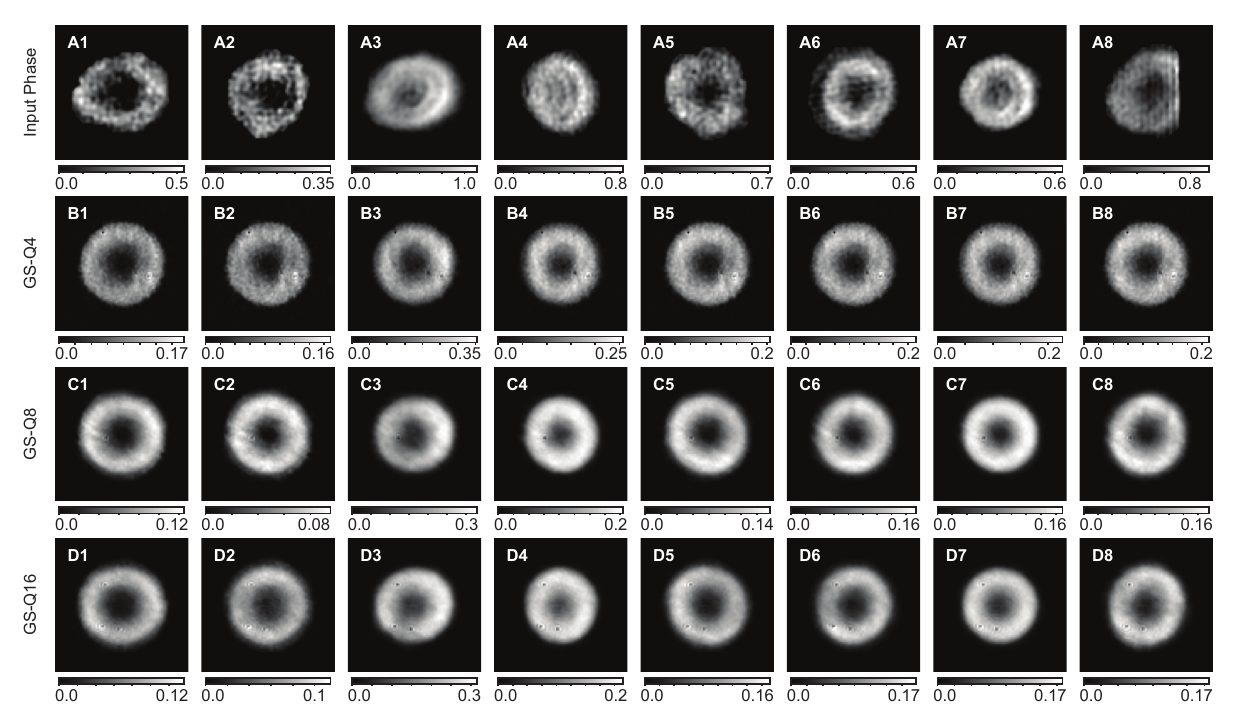}
    \caption{\textbf{Mode collapse of D2NNs trained using GS for RBC dataset:} \textbf{A1)-A8)} show eight randomly selected examples from the RBC test set. Subsequent rows show the resulting output intensities from D2NNs trained using GS with 4-level (\textbf{B1-B8}), 8-level (\textbf{C1-C8}), and 16-level (\textbf{D1-D8}) quantized phase weights. Although the inputs have different morphologies, each D2NN gives similar outputs to all the inputs.}
    \label{fig:mode_collapse_sup}
\end{figure*}

\begin{figure*}
        \centering
        \includegraphics[width=\linewidth]{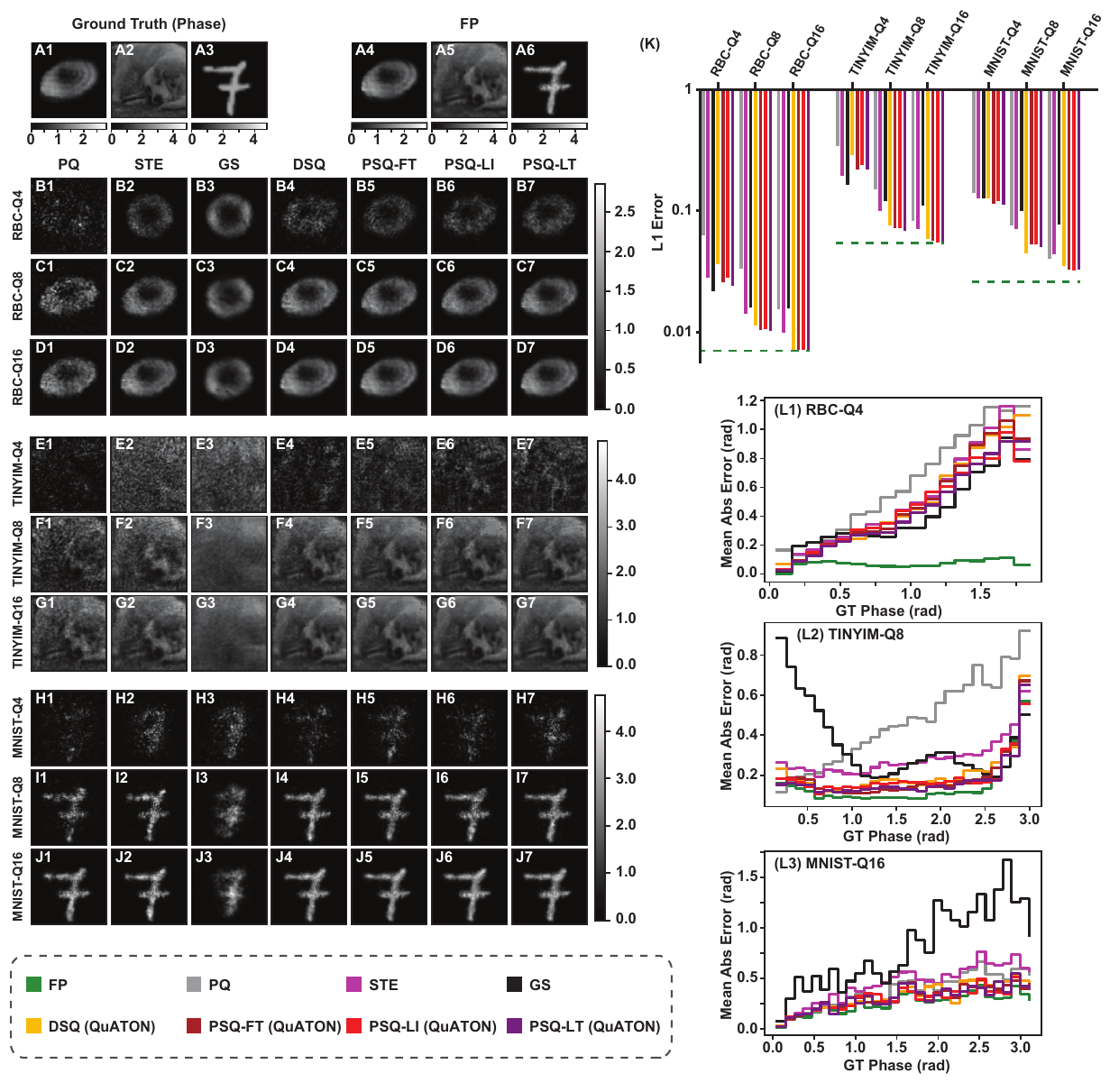}
        \caption{\textbf{All-optical quantitative phase imaging results (additional examples):} \textbf{A1)-A3)} show three examples of the phase of the incoming wave to the D2NN for the three datasets considered. \textbf{A4-A6)} show the output intensities $\times \pi$ for the D2NNs trained with full precision (FP) weights. Rows \textbf{B)-J)} show the QPI results for quantization-aware trained D2NNs for each of the examples. Each row named as \textbf{\textit{x}-Q\textit{n}} shows the results for dataset \textbf{\textit{x}} $\in \{\textrm{RBC, TINYIM, MNIST}\}$, using D2NNs trained with \textbf{\textit{n}}-level quantized weights ($\textbf{\textit{n}} \in \{4, 8, 16\}$). Each column corresponds to different QAT methods considered which are stated above the row B). Note that all the results are given as $\textrm{output intensity} \times \pi$. \textbf{K)} shows the comparison of the mean L1 error of the predictions over the test set of each dataset. \textbf{L1)-L3)} show mean absolute phase error variation against ground truth phase for \textbf{RBC-Q4, TINYIM-Q8,} and \textbf{MNIST-Q16} cases respectively. These plots are shown for the given examples in the figure.}
        \label{fig:qpi_sup}
    \end{figure*}

\begin{figure*}
    \centering
    \includegraphics[width=0.8\linewidth]{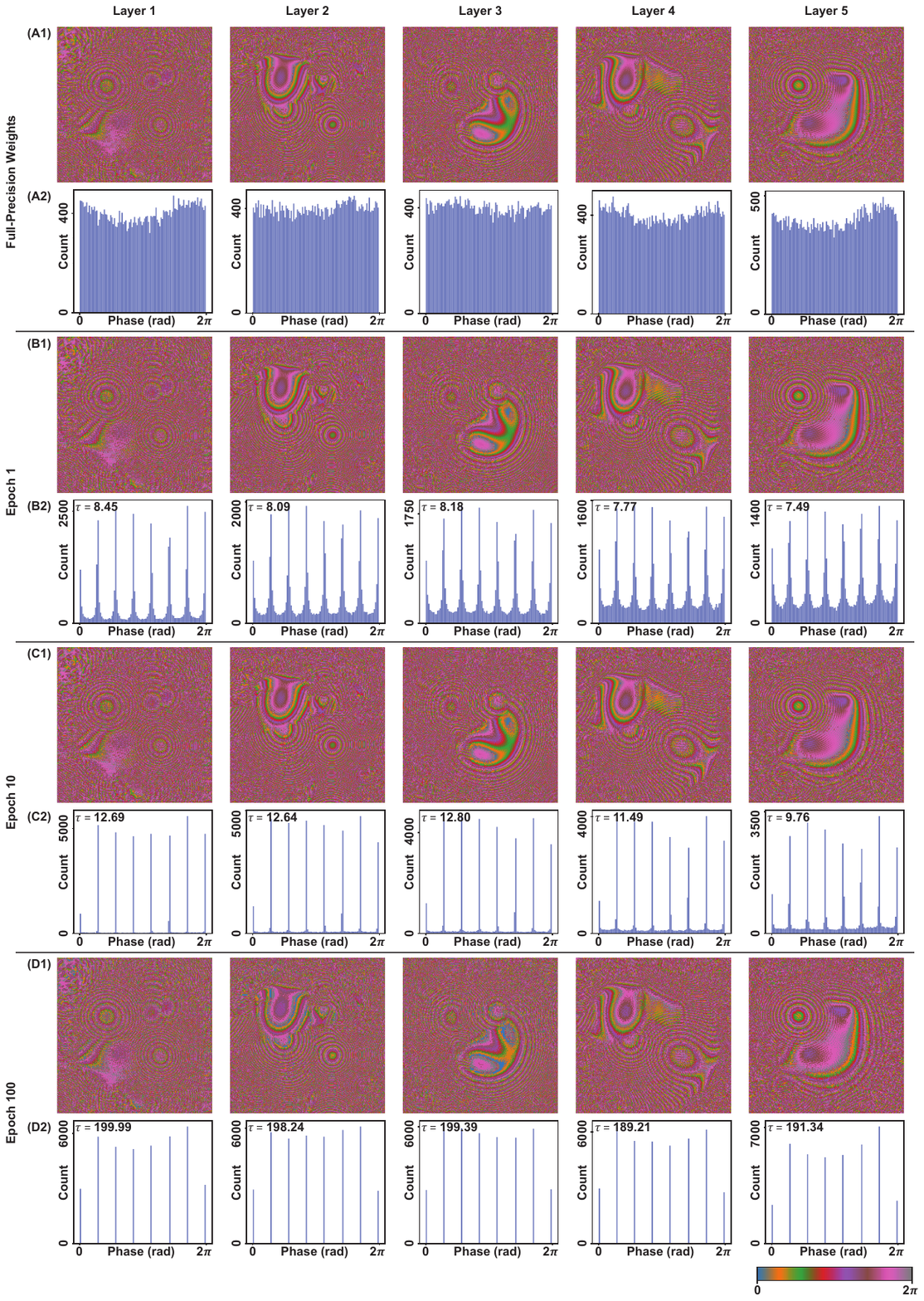}
    \caption{\textbf{Progressive training for quantization (For TINYIM-Q8 case using PSQ-LT method):} \textbf{A1)} and \textbf{A2)} shows full-precision initialization and the weight distributions of the phase maps for each D2NN layer. Rows \textbf{B)} - \textbf{D)} shows the phase maps and weight distributions for epochs 1, 10, and 100 respectively during progressive training. In each epoch, the learned temperature factor for each layer is shown in the top-left corner of the histograms.}
    \label{fig:progressive_training}
\end{figure*}

\end{document}